\documentclass[conference]{IEEEtran}
\usepackage{cite}

\ifCLASSINFOpdf
   \usepackage[pdftex]{graphicx}
   \graphicspath{{Figures/}}
\else
   \usepackage[dvips]{graphicx}
   \graphicspath{{Figures/}}
\fi
\usepackage{amsmath}
\usepackage{algorithmic}

\usepackage{array}

\ifCLASSOPTIONcompsoc
 \usepackage[caption=false,font=normalsize,labelfont=sf,textfont=sf]{subfig}
\else
 \usepackage[caption=false,font=footnotesize]{subfig}
\fi
\usepackage{fixltx2e}
\usepackage{url}

\usepackage{hyperref}
\usepackage{ragged2e}
\usepackage{amsfonts,amssymb}
\usepackage{multirow}
\usepackage{listings}
\usepackage{xcolor}
\usepackage{braket}
\usepackage{booktabs}

\usepackage{upgreek}
\newcommand{\ucaspian}{$\upmu$Caspian}

\usepackage{soul}

\IEEEoverridecommandlockouts

\begin{document}

\lstset{
    breaklines=true,
    prebreak=\raisebox{0ex}[0ex][0ex]{\ensuremath{\hookleftarrow}},
    tabsize=1,
    basicstyle=\ttfamily
}

%
\title{Encoding Integers and Rationals on Neuromorphic Computers using Virtual Neuron\\
}


\author{
\IEEEauthorblockN{Prasanna Date}
\IEEEauthorblockA{Oak Ridge National Laboratory\\
Oak Ridge, Tennessee 37830\\
Email: datepa@ornl.gov} \\

\IEEEauthorblockN{Catherine Schuman}
\IEEEauthorblockA{University of Tennessee\\
Knoxville, Tennessee 37996\\
Email: cschuman@utk.edu} \\

\and

\IEEEauthorblockN{Shruti Kulkarni}
\IEEEauthorblockA{Oak Ridge National Laboratory\\
Oak Ridge, Tennessee 37830\\
Email: kulkarnisr@ornl.gov} \\

\IEEEauthorblockN{Thomas Potok}
\IEEEauthorblockA{Oak Ridge National Laboratory\\
Oak Ridge, Tennessee 37830\\
Email: potokte@ornl.gov} \\

\and

\IEEEauthorblockN{Aaron Young}
\IEEEauthorblockA{Oak Ridge National Laboratory\\
Oak Ridge, Tennessee 37830\\
Email: youngar@ornl.gov} \\

\IEEEauthorblockN{Jeffrey Vetter}
\IEEEauthorblockA{Oak Ridge National Laboratory\\
Oak Ridge, Tennessee 37830\\
Email: vetter@ornl.gov}

}


%


\maketitle

\begin{abstract}
Neuromorphic computers perform computations by emulating the human brain, and use extremely low power. They are expected to be indispensable for energy-efficient computing in the future. While they are primarily used in spiking neural network-based machine learning applications, neuromorphic computers are known to be Turing-complete, and thus, capable of general-purpose computation. However, to fully realize their potential for general-purpose, energy-efficient computing, it is important to devise efficient mechanisms for encoding numbers. Current encoding approaches have limited applicability and may not be suitable for general-purpose computation. In this paper, we present the virtual neuron as an encoding mechanism for integers and rational numbers. We evaluate the performance of the virtual neuron on physical and simulated neuromorphic hardware and show that it can perform an addition operation using $23$ nJ of energy on average using a mixed-signal memristor-based neuromorphic processor. We also demonstrate its utility by using it in  some of the $\mu$-recursive functions, which are the building blocks of general-purpose computation.
\end{abstract}


%
\IEEEpeerreviewmaketitle

\section{Introduction}
\label{sec:intro}

Neuromorphic computers perform computations by emulating the human brain \cite{calimera2013human}.
Akin to the human brain, they are extremely energy efficient in performing computations \cite{grollier2020neuromorphic}.
For instance, while CPUs and GPUs consume around 70 W and 250 W of power, a neuromorphic computer consumes around 65 mW of power, i.e. 4--5 orders of magnitude less power than CPUs and GPUs \cite{akopyan2015truenorth}.
The structural and functional units of neuromorphic computation are neurons and synapses, which can be implemented on digital or analog hardware \cite{schuman2020resilience}.
They impart critical characteristics to neuromorphic computing such as co-located processing and memory, event-driven computation, massively parallel operation and inherent scalability \cite{schuman2022opportunities}.
These characteristics are crucial for the energy efficiency of neuromorphic computers.
For the purposes of this paper, we define neuromorphic computing as any computing paradigm (theoretical, simulated, or hardware) that performs computations by emulating the human brain, i.e., by using neurons and synapses, that communicate with binary-valued signals (also known as spikes).

Neuromorphic computing is primarily used in machine learning applications, almost exclusively leveraging spiking neural networks (SNN) \cite{ghosh2009spiking}.
In the recent years however, it has also been used in non-machine learning applications such as graph algorithms, boolean linear algebra and neuromorphic simulations \cite{kay2020neuromorphic,schuman2021sparse,hamilton2020modeling}.
Researchers have also shown that neuromorphic computing is Turing-complete, i.e. capable of general-purpose computation \cite{date2021neuromorphic}.
This ability to perform general-purpose computations and potentially use orders of magnitude less energy in doing so is why neuromorphic computing is poised to be an indispensable part of the energy-efficient computing landscape in the future.
However, in order to realize a fully operational, general-purpose neuromorphic computer, we must address several limitations of today's neuromorphic computing at the hardware and software level.

One of the biggest limitations of neuromorphic computing today is the inability to encode numbers efficiently \cite{huynh2022implementing}. While there are several studies on the performance of neural network models with low precision representation of parameters such as weights \cite{kang2017dynamic}, these approximate representations are not suitable for general purpose computing.
There exist several methods to encode numbers on neuromorphic computers \cite{schuman2019non}.
However, their scope is restricted to the specific application for which they were designed and is not suitable for general-purpose computation.
Furthermore, no good mechanism exists for encoding negative integers and positive and negative rational numbers exactly on neuromorphic computers.
The ability to encode basic data types such as numbers, letters and symbols is vital for any computing platform.
Efficient mechanisms for encoding rational numbers would significantly expand the scope of neuromorphic computing to new application areas such as non-SNN-based machine learning (regression, support vector machines etc.), wide range of graph and network problems, general-purpose computing applications, linear and non-linear optimization, simulation of physical systems and perhaps even finding good solutions to NP-complete problems.
Working with rational numbers is central to these application areas.
Being able to encode rational numbers on a neuromorphic computer could enable us to address these problems in an energy-efficient manner.

Neuromorphic computers are seen as accelerators that can perform machine learning tasks using spiking neural networks.
In order to perform any other operation (for example, arithmetic, logical, relational, etc.), we still resort to CPUs and GPUs because no good neuromorphic methods exist to perform these operations.
To perform these operations, we have to transfer data from the neuromorphic computer to CPUs/GPUs, which incurs significant communication costs (more than $99\%$ of the time is spent in communication) and is highly inefficient.
Devising neuromorphic approaches for performing these non-machine learning operations would drastically reduce the cost of transferring data to and from the neuromorphic computer.
This would enable performing all types of computation (machine learning as well as non-machine learning) efficiently on low power neuromorphic computers deployed on the edge.

To this extent, we present the virtual neuron for addressing the limitation of neuromorphic computers to encode numbers.
This is the first step towards performing general-purpose computations on neuromorphic computers.
To the best of our knowledge, the virtual neuron is the first encoding mechanism that can encode positive and negative integers and rational numbers on a neuromorphic computer.
Specifically, our main contributions are:
\begin{enumerate}
    \item We introduce the virtual neuron, which is made up of spiking neurons and synapses. It is a spatial encoding mechanism that leverages the binary representation of numbers to encode positive and negative integers and rational numbers. This is detailed in Section \ref{sec:virtual-neuron}.
    \item We implement the virtual neuron in the Neural Simulation Technology (NEST) simulator \cite{Gewaltig:NEST} and test its performance on 8, 16 and 32 bit rational numbers. We also discuss the computational complexity of the virtual neuron. This is outlined in Sections \ref{sec:implementation-details} and \ref{sec:testing-results}.
    \item We analyze the performance of the virtual neuron on neuromorphic hardware by analyzing the run time using a digital neuromorphic hardware design, and we also estimate the energy usage of the virtual neuron on a mixed-signal memristor-based hardware design.
    \item We demonstrate the usability of the virtual neuron by using it in five functions: constant function, successor function, predecessor function, multiply by $-1$ function and $N$-neuron addition. This is covered in Section \ref{sec:applications}. Without the use of virtual neuron, implementing these functions on a neuromorphic computer would be extremely challenging.
\end{enumerate}

\section{Related Work}
\label{sec:related-work}

Neuromorphic computing was introduced by Carver Mead in the 1980s \cite{mead1990neuromorphic}.
Since then, it is primarily used for SNN-based machine learning applications, including computer vision \cite{serre2010neuromorphic}, natural language \cite{sung2021memory} and speech recognition \cite{blouw2020event}.
These applications are mainly found in embedded systems, edge computing and Internet of Things (IoT) settings as they have strict requirements for small size, weight and power \cite{liu2020quantized,covi2021adaptive,fayyazi2018ultra}.
Several on-chip as well as off-chip learning algorithms that leverage gradient-based as well as local learning rules have been suggested for training SNNs in neuromorphic applications \cite{tavanaei2019deep,date2019combinatorial,lee2016training,mohemmed2013training}.
Neuromorphic computing has also been used in neuroscience simulations \cite{indiveri2021introducing}.
These simulations span a wide range of neuron and synapse models, the most popular of which is the leaky-integrate-and-fire (LIF) neuron model \cite{burkitt2006review}.
Our virtual neuron will use spiking neurons that are of the LIF type as well.
The latest additions to the arsenal of neuromorphic computing applications include graph algorithms \cite{kay2020neuromorphic,kay2021neuromorphic,hamilton2020spike}, autonomous racing \cite{patton2021neuromorphic}, epidemiological simulations \cite{hamilton2020modeling}, classifying supercomputer failures \cite{date2018efficient}, $\mu$-recursive functions \cite{date2021neuromorphic}, and boolean matrix-vector multiplication \cite{schuman2021sparse}.
With regards to designing neuromorphic algorithms, a theoretical framework for determining the computational complexity has also been proposed \cite{date2021computational}.

Most of the above applications are based on binary numbers and Boolean arithmetic.
This is largely due to the spiking behavior of the neuron---the spikes can be interpreted as a $1$, whereas lack of spike can be interpreted as a $0$.
This spiking behavior naturally lends itself to binary or Boolean operations.
Leveraging this behavior, several mechanisms for encoding numbers (mainly positive integers) have been proposed in the literature.
Choi et al. propose a neuromorphic implementation of hypercolumns, including mechanisms for encoding images \cite{choi2005neuromorphic}.
Cohen et al. use neuromorphic methods to classify images that have been encoded as spikes \cite{cohen2016skimming}.
Hejda et al. present a mechanism for encoding image pixels as rate-coded optical spike trains \cite{hejda2021neuromorphic}.
Sengupta and Roy encode neural and synaptic functionalities in electron spin as an efficient way to perform neuromorphic computation \cite{sengupta2017encoding}.
Yi et al. propose a field programmable gate array (FPGA) platform to be used as a spike time dependent encoder and dynamic reservoir in neuromorphic computers \cite{yi2016fpga}.

Iaroshenko and Sornborger propose neuromorphic mechanisms for encoding binary numbers, and use it for binary two's complement operations and binary matrix multiplication \cite{iaroshenko2021binary}.
However, their approach uses numbers of neurons and synapses that are of the quadratic and cubic order respectively.
Lawrence et al. perform neuromorphic matrix multiplication by using an intermediate transformation matrix for encoding that is flattened into a neural node \cite{lawrence2021matrix}.
Schuman et al. propose three ways of encoding positive integers on neuromorphic computers, which are in turn used in many different applications \cite{schuman2019non}.
Zhao et al. develop a compact, low power, and robust spiking-time-dependent encoder, designed with a LIF neuron cluster and a chaotic circuit with ring oscillators \cite{zhao2015neuromorphic}.
Zhao et al. develop a method for representing data using spike time dependent encoding that efficiently maps a signal's amplitude to a spike time sequence representing the input data \cite{zhao2015spike}.
Zhao et al. propose an analog temporal encoder for making neuromorphic computing robust and energy efficient \cite{zhao2016making}.
Wang et al., made use of radix encoding of spike to realize SNNs more efficiently and improve the speedup by reducing the overall latency for machine learning applications  \cite{wang2021efficient}. Other efforts to realize basic computations on neuromorphic platforms leveraging the inherent structure and parameters of SNNs for logic operations such as AND, OR and XOR have been demonstrated in \cite{plank2021spiking}.
George et al., performed IEEE 754 compliant addition using SNNs by designing a system based on the Neural Engineering Framework (NEF) and implemented, simulated, and tested the design using Nengo \cite{George2019}. This approach uses an ensemble of 300 neurons to represent each bit and the function of each component in the adder is approximated using NEF to determine the appropriate synapse weights. Dubey et al., extend this work to perform IEEE 754 compliant multiplication using the same encoding method and similar methodology of using NEF to approximate the functions of the multiplier sub-components\cite{Dubey2020}. 

Most of these encoding mechanisms have the ability to encode binary or Boolean numbers, with some being able to encode positive integers as well.
These methods are designed with specific applications in mind such as image applications, and it is not clear if they can be used for general-purpose neuromorphic computation, where arithmetic operations need to be performed on positive and negative integers/rationals.
Moreover, some of the encoding mechanisms such as binning tend to lose information by virtue of discretization.
To the best of our knowledge, an efficient mechanism for encoding positive and negative rational numbers does not exist in the neuromorphic literature yet.
We address this gap by proposing the virtual neuron.
In our quest for general-purpose, energy-efficient neuromorphic computing, being able to encode rational numbers is a critical milestone.

\section{Neuromorphic Computing Model}
\label{sec:model}

Neuromorphic computing systems implement vastly different neuron and synapse models, and the precise model details depend on the specific hardware implementation.
We leverage the neuromorphic computing model described in \cite{date2021neuromorphic} and \cite{date2021computational}, which is based on the LIF neuron with two parameters (threshold, $\nu$ and leak $\lambda$), and two synapse parameters (weight, $\omega$ and delay, $\delta$).

\section{The Virtual Neuron}
\label{sec:virtual-neuron}

\begin{figure*}[t!]
\centering
\subfloat[Encoding $10$]{
    \includegraphics[trim={300 90 300 100},clip,scale=0.45,page=1]{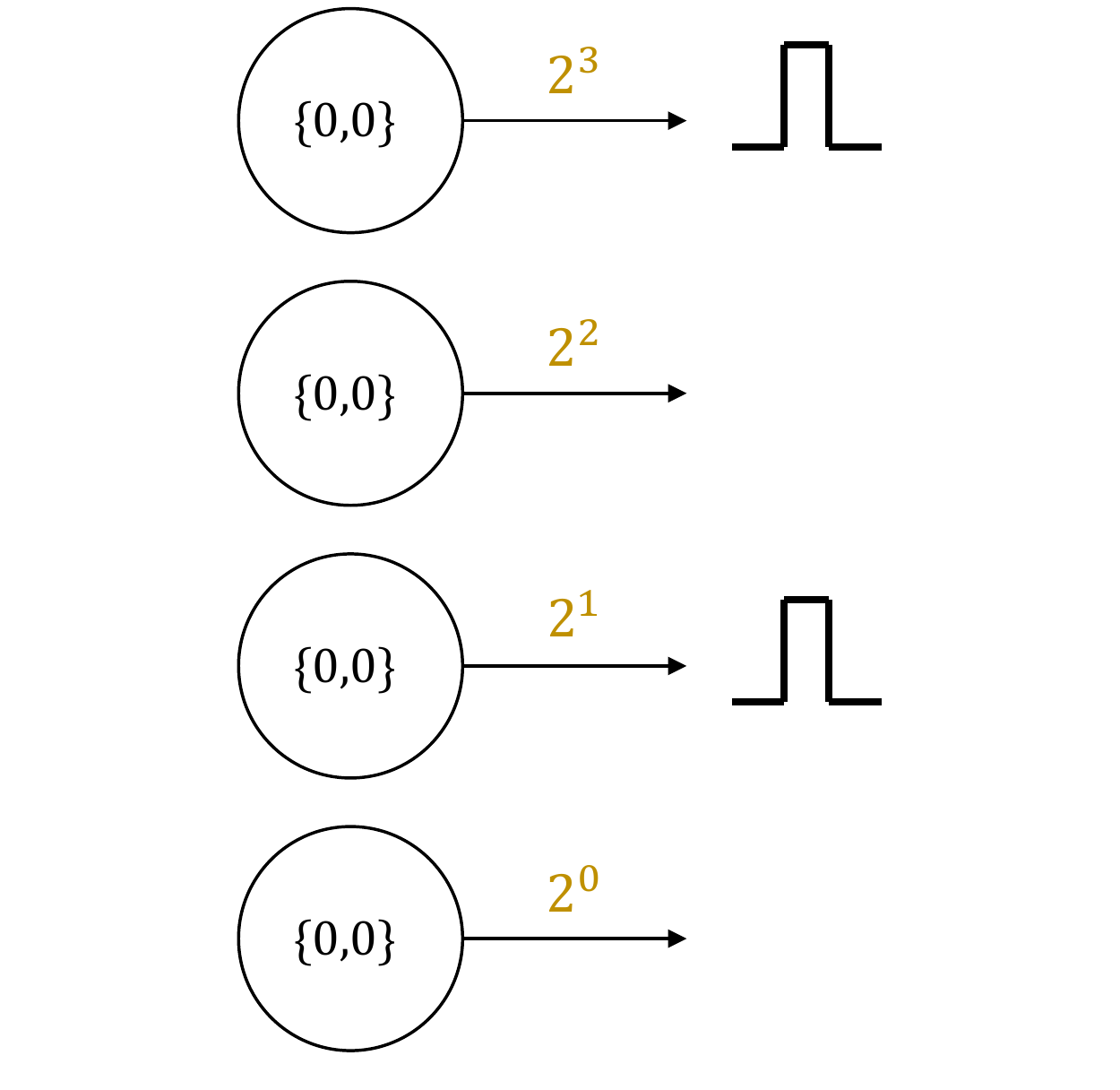}
    \label{subfig:positive-integer-encoding}
}
\subfloat[Encoding $0.625$]{
    \includegraphics[trim={300 90 300 100},clip,scale=0.45,page=2]{virtual-neuron-circuits.pdf}
    \label{subfig:positive-rational-encoding}
}
\subfloat[Encoding $-3.5$]{
    \includegraphics[trim={300 90 300 100},clip,scale=0.45,page=3]{virtual-neuron-circuits.pdf}
    \label{subfig:negative-rational-encoding}
}%
\caption{Encoding mechanism of the virtual neuron. Numbers can be encoded by selecting appropriate synaptic weights. Here, we use four neurons to encode: (\ref{subfig:positive-integer-encoding}) positive integers; (\ref{subfig:positive-rational-encoding}) positive rationals; and, (\ref{subfig:negative-rational-encoding}) negative rationals. The four spiking neurons can encode four bits of information.}
\label{fig:encoding-mechanism}
\end{figure*}

\begin{figure}[t!]
    \centering
    \includegraphics[trim={200 40 100 40},clip,scale=0.4,page=4]{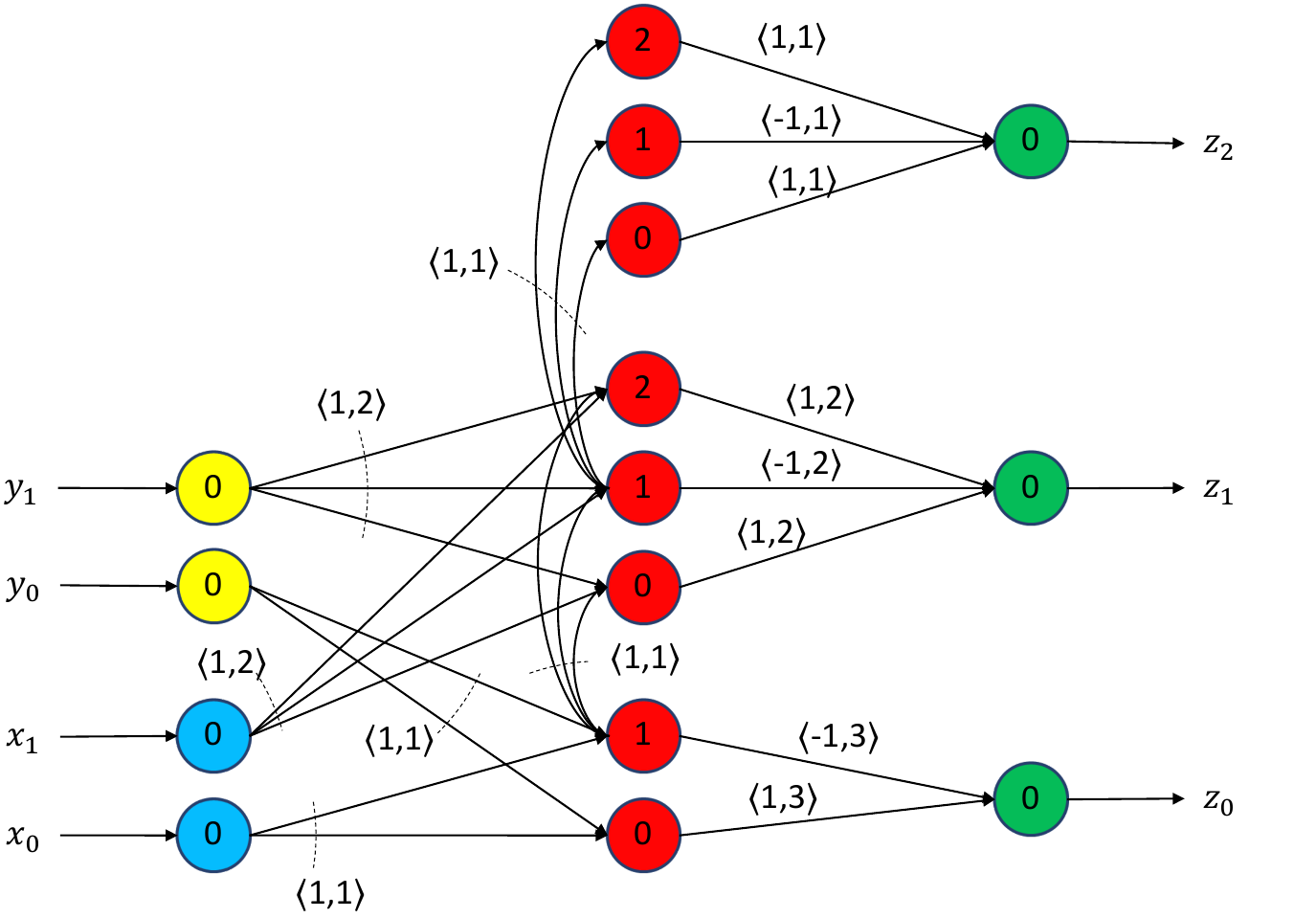}
    \caption{Two-bit virtual neuron. Takes two 2-bit numbers as input on the left: $X$ and $Y$, represented as $x_1$, $x_0$ and $y_1$, $y_0$ respectively. Adds the two numbers and generates their sum on the right. The sum of two 2-bit numbers can at most be a 3-bit number.}
    \label{fig:2-bit-virtual-neuron}
\end{figure}

Structurally, the virtual neuron is composed of a group of LIF neurons and synapses that are connected in a particular way.
Functionally, the virtual neuron mimics the behavior of an artificial neuron with identity activation.
The virtual neuron is an encoding mechanism as well as an adder.
It performs the addition operation similar to a ripple carry adder.
The rationale behind the encoding mechanism of the virtual neuron is rooted in the binary encoding of numbers.
Figure \ref{fig:encoding-mechanism} shows three ways of encoding four bit numbers on a neuromorphic computer.
Notice that each neuron in the figure represents a bit.
The synapse coming out of the neuron assigns a value to the binary spike of the neuron by multiplying it with its synaptic weight.
By having powers of two as the synaptic weights, we can encode rational numbers using a group of neurons.
For instance, the synapses coming out of the four neurons in Figure \ref{subfig:positive-integer-encoding} have weights $2^0$, $2^1$, $2^2$ and $2^3$.
When the second and fourth neurons (from the bottom) spike, the result gets multiplied by $2$ and $8$ in the outgoing synapses respectively.
This is interpreted as the number $10$ under this encoding mechanism.
Similarly, we can set the synaptic weights to be negative powers of two as shown in Figure \ref{subfig:positive-rational-encoding}.
This enables us to encode positive fractions as well.
When the first and third neurons (from the bottom) spike as shown in the figure, the result is interpreted as a $0.625$.
Lastly, if the synaptic weights are set to negatives of positive and negative powers of two as shown in Figure \ref{subfig:negative-rational-encoding}, we can encode negative rational numbers.
When the three neurons spike in the figure, the output is interpreted as $-3.5$.

We now show how the virtual neuron can integrate the incoming signals and generate a rational number as output.
For ease of explanation, we stick to the two-bit virtual neuron as shown in Figure \ref{fig:2-bit-virtual-neuron}.
The two-bit virtual neuron takes as input two 2-bit numbers $X$ and $Y$, shown in the figure as $[x_1, x_0]$ (blue neurons) and $[y_1, y_0]$ (yellow neurons) respectively.
It then adds X and Y in the three groups of bit neurons, which are shown in red.
We call them bit neurons because they are responsible for the bit-level operations in the circuit such as bitwise addition, propagating the carry bit etc.
Finally, it produces a 3-bit number $Z$ as output, shown in the figure as $[z_2, z_1, z_0]$ (green neurons).

The default internal states of all neurons are set to $-1$.
Furthermore, all neurons have a leak of $0$, which means they reset to their default internal state instantaneously if they do not spike.
The reset state (or reset voltage) of all neurons is set to $-1$, so that the internal state of all neurons will be reset to $-1$ after they spike.
The numbers on the neurons indicate their thresholds, for e.g., the top set of bit neurons (red neurons) have thresholds $0$, $1$ and $2$ respectively.
The synapse parameters are indicated in angular brackets on top or bottom of the synapses.
The first parameter is the synaptic weight and the second parameter is the synaptic delay.
If a group of synapses has the same parameters, it is indicated using a dotted arc.
The synaptic delays are adjusted such that the bit operations of red neurons are synchronized and the output $Z$ is produced at the same time.

We now go over the inner workings of the virtual neuron shown in Figure \ref{fig:2-bit-virtual-neuron} by taking the example: $[x_1, x_0] = [1, 1]$, and $[y_1, y_0] = [0, 1]$.
We start our analysis when the inputs $X$ and $Y$ have been received in the blue and yellow neurons---let us call this the zeroth time step.
In the first time step, the bottom set of bit neurons in red receive an input of $1$ along each of their incoming synapses.
Thus, the total incoming signal at both these neurons is $2$, which changes their internal state from $-1$ to $1$.
As a result, both the bottom red neurons spike.
Their spikes are sent along their outgoing synapses, which delay the signal for $3$ time steps.

In the second time step, the middle group of bit neurons receive all of their inputs: $1$ from the blue incoming neuron representing $x_1$, $0$ from the yellow neuron representing $y_1$ and $1$ from the bit neuron with a threshold of $1$ in the bottom group.
Thus, sum of their incoming signals is $2$ and their internal states reach a value of $1$.
As a result, neurons with thresholds $0$ and $1$ in the middle group of bit neurons spike, whereas the one with threshold $2$ does not spike.
The spikes from the middle red neurons with thresholds $0$ and $1$ are sent to the green output neuron representing $z_1$ along their outgoing synapses, which stall for $2$ time steps.

In the third time step, the three bit neurons in the top group of red neurons receive an input of $1$ along each of their incoming synapses.
As a result, their internal states are incremented by $1$ to the value of $0$.
The neuron with $0$ threshold spikes as a result and sends its spike along its outgoing synapse to the green neuron representing $z_2$.

In the fourth time step, the green neurons representing $z_0$, $z_1$ and $z_2$ receive their inputs.
$z_0$ receives a $1$ and $-1$ from the bit neurons with the thresholds $0$ and $1$ respectively in the bottom group of red neurons.
Its total input is thus $1 - 1 = 0$, which keeps its internal state at $-1$, and it does not spike.
Similar operations happen at the green neuron representing $z_1$.
It too does not spike.
The green neuron representing $z_2$ receives a signal of $1$ from the bit neuron with the threshold of $0$ in the top red set.
As a result, its internal state is incremented by $1$ to the value of $0$, and it spikes.

The net output $[z_2, z_1, z_0]$ from the circuit is $[1, 0, 0]$, which can be interpreted as a $4$ in binary.
Given that our inputs were $[x_1, x_0] = [1, 1]$, and $[y_1, y_0] = [0, 1]$, i.e., $X = 3$, and $Y = 1$, we have received the correct output of $4$ from the virtual neuron circuit.
While we restricted ourselves to 2-bit positive integers in this example, we show in the subsequent subsections that similar circuits can be used to encode and add two rational numbers in the virtual neuron and generate a rational number as output.
Finally, note that we did not use powers of two in the synapses inside of the virtual neuron.
Depending on the application, powers of two as synaptic weights may be used on the incoming or outgoing synapses for a given virtual neuron.

In the following subsections, we present virtual neuron circuits that have higher precision.
We let $P_+$ and $P_-$ denote the number of bits used to represent positive and negative numbers respectively.
We call them positive precision and negative precision respectively.
In general, the positive precision $P_+$ will be distributed among bits used to represent positive integers ($2^0, 2^1, 2^2, \ldots$) and positive fractionals ($2^{-1}, 2^{-2}, 2^{-3}, \ldots$).
Similarly, the negative precision $P_-$ will be distributed among bits used to represent negative integers ($-2^0, -2^1, -2^2, \ldots$) and negative fractionals ($-2^{-1}, -2^{-2}, -2^{-3}, \ldots$).

We now describe the connections for a virtual neuron with arbitrary precision.
Each input neuron has both threshold and leak as zero.
Each input $x_i$ and $y_i$ is connected to the set of bit neurons corresponding to bit $i$.
In the case of bit $0$ there are two such bit neurons, while for every other bit, there are three neurons per bit shown in red.
The synaptic weights of all these connections are unity and their delays are $i+1$.
Each set of bit neurons has neurons with thresholds of zero and one.
All bit neurons except the zero bit have a neuron with a threshold of two as well.
The neuron with a threshold of one in the set of neurons representing bit $i$ is connected to all neurons in the $(i+1)$-th set.
This neuron is responsible for the propagating the carry bit to the next set of bit neurons.
It spikes only when there is a carry operation to be performed at the $i$-th bit.
The carry synapses have both weights and delays as unity.
The bit neurons of the $i$-th bit are connected to the $i$-th output neuron.
The synaptic weights for the bit neurons having thresholds of zero and two are $1$, while those for the bit neurons having threshold of one are $-1$.
The $-1$ weight is seen as an inhibitory connection that cancels the signal coming from the neuron with threshold zero in the same bit set.
The delays on the synapses going from $i$-th bit set to the $i$-th output neuron are set to $\max\{P_+, P_-\} - i + 1$.
This delay ensures that all output neurons spike at the same time.

\subsection{Positive Integers}
\label{sub:virtual-neuron/positive-integers}

\begin{figure}[t!]
    \centering
    \includegraphics[trim={100 0 200 0},clip,scale=0.4,page=5]{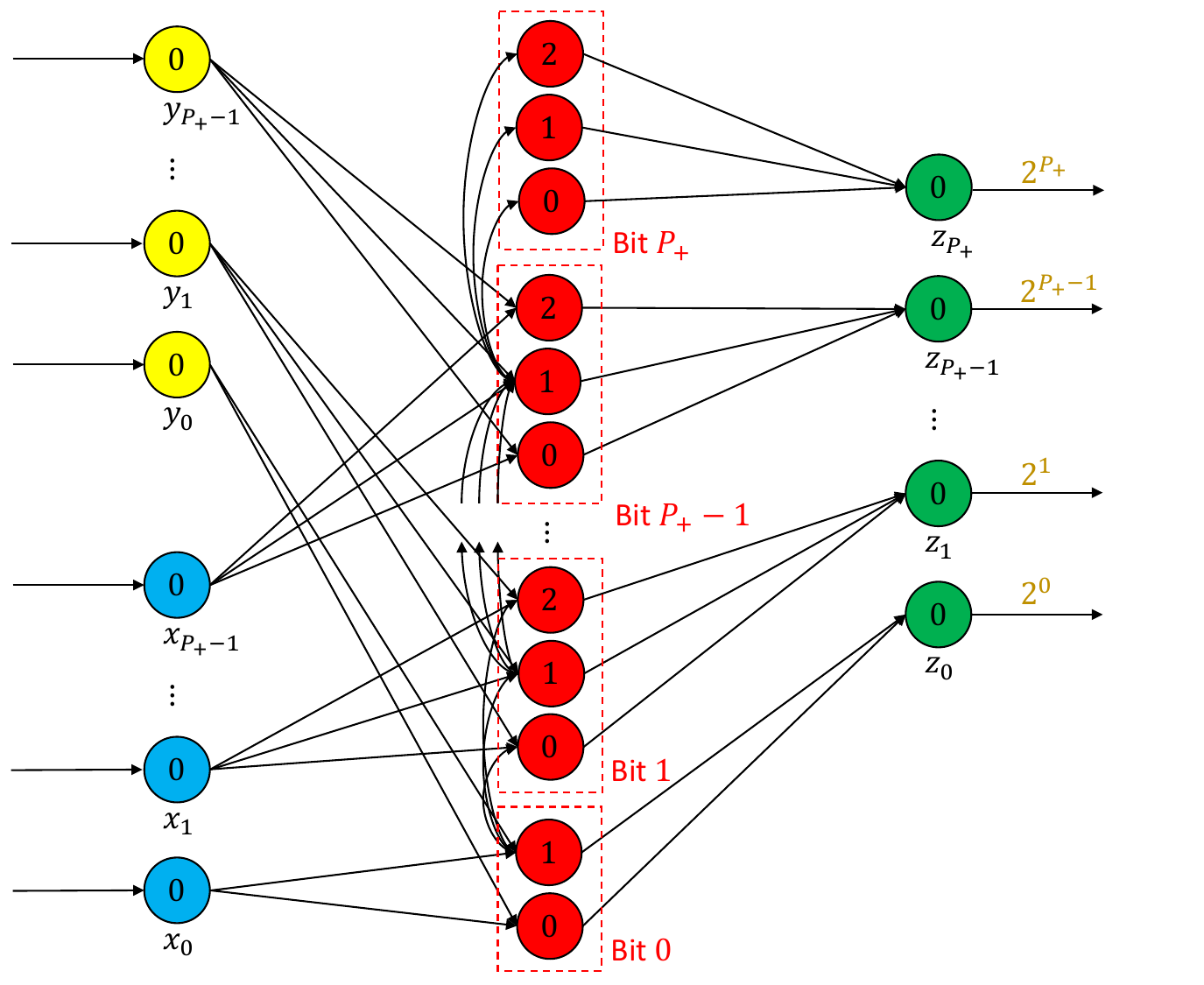}
    \caption{$P_+$ bit virtual neuron for encoding positive integers. Synapse parameters are omitted for brevity. Actual synapse parameters are assumed to be similar to the circuit shown in Figure \ref{fig:2-bit-virtual-neuron}.}
    \label{fig:positive-integers}
\end{figure}

Figure \ref{fig:positive-integers} shows the virtual neuron circuit that takes two $P_+$ bit numbers $X$ and $Y$ as inputs, shown as blue and yellow neurons respectively.
The bit-level addition and carry operations are performed by the bit neurons shown in red.
There are $P_+ + 1$ groups of these bit neurons.
Finally, the output of the virtual neuron $Z$ has $P_+ + 1$ bit precision, and is shown by the green output neurons.
In the figure, we omit synapse parameters for brevity.
Notice that the synaptic weights on the outgoing synapses are positive powers of $2$.

\subsection{Positive Fractionals}
\label{sub:virtual-neuron/positive-fractionals}

\begin{figure}[t!]
    \centering
    \includegraphics[trim={100 0 200 0},clip,scale=0.4,page=6]{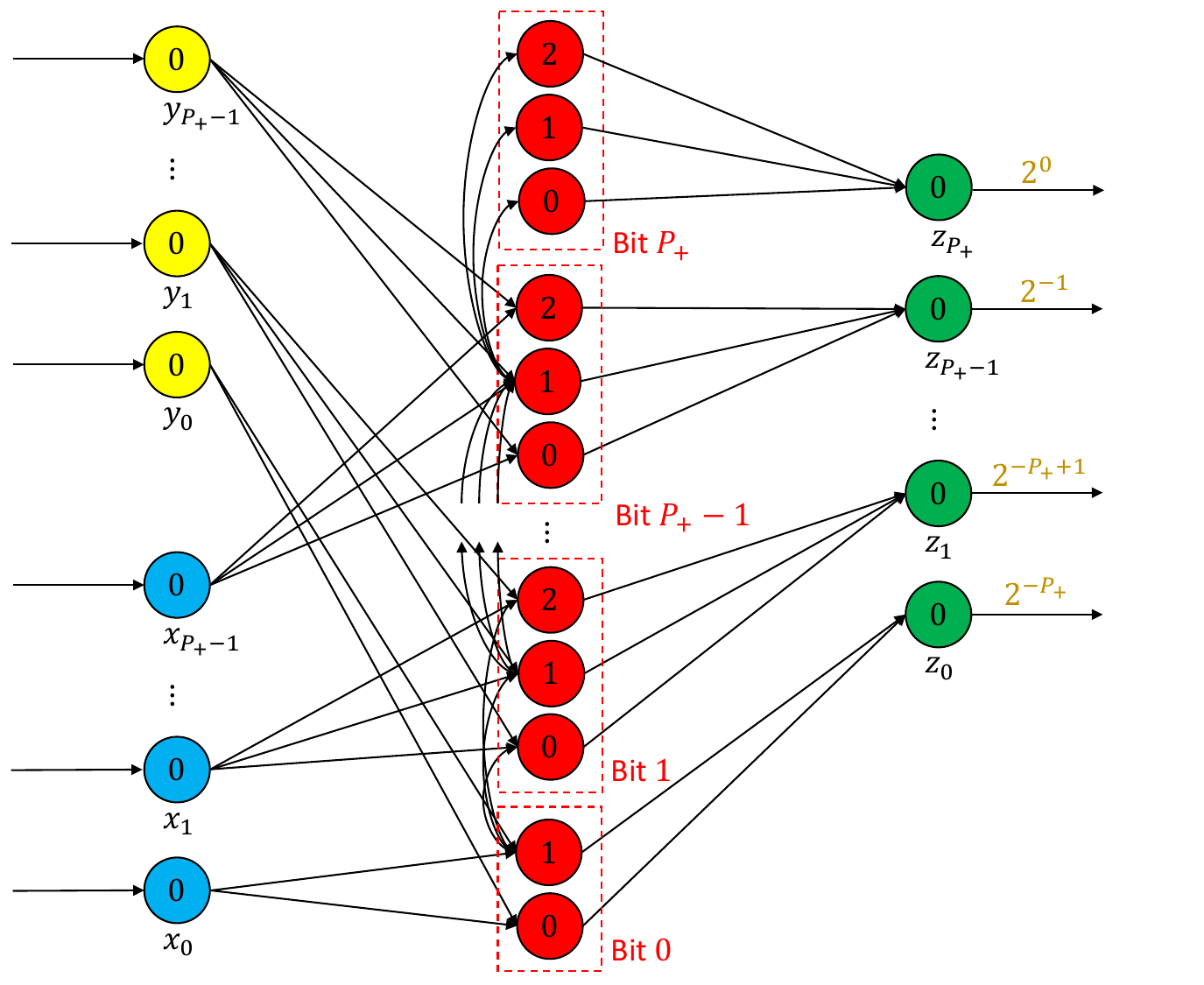}
    \caption{$P_+$ bit virtual neuron for encoding positive fractionals. Synapse parameters are omitted for brevity. Actual synapse parameters are assumed to be similar to the circuit shown in Figure \ref{fig:2-bit-virtual-neuron}.}
    \label{fig:positive-fractionals}
\end{figure}

Figure \ref{fig:positive-fractionals} shows the $P_+$ bit virtual neuron for encoding positive fractionals.
The circuit is almost identical to Figure \ref{fig:positive-integers}.
The only difference is in the synaptic weights of the outgoing synapses.
In this case, these synapses have negative powers of two, i.e., $2^0, 2^{-1}, 2^{-2}, 2^{-3}, \ldots$ as their weights.

\subsection{Negative Integers}
\label{sub:virtual-neuron/negative-integers}

\begin{figure}[t!]
    \centering
    \includegraphics[trim={100 0 200 0},clip,scale=0.4,page=7]{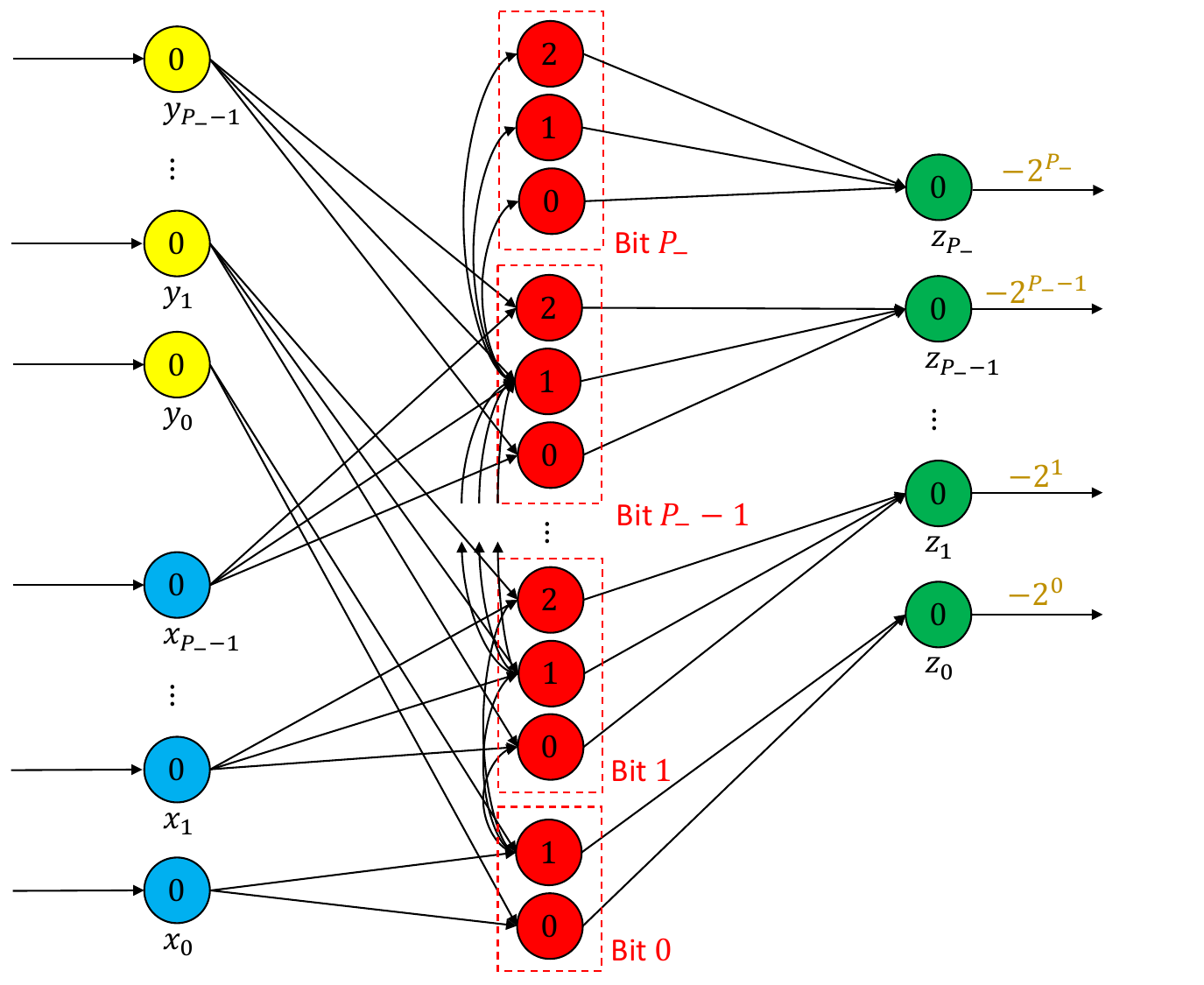}
    \caption{$P_-$ bit virtual neuron for encoding negative integers. Synapse parameters are omitted for brevity. Actual synapse parameters are assumed to be similar to the circuit shown in Figure \ref{fig:2-bit-virtual-neuron}.}
    \label{fig:negative-integers}
\end{figure}

Figure \ref{fig:negative-integers} shows the virtual neuron circuit for encoding negative integers.
It takes two $P_-$ bit numbers $X$ and $Y$ as inputs.
After standard virtual neuron operations, a $P_- + 1$ bit number $Z$ is produced as the output.
In this case, these weights are negatives of positive powers of two, i.e., $-2^0, -2^1, -2^2, \ldots$.

\subsection{Negative Fractionals}
\label{sub:virtual-neuron/negative-fractionals}

\begin{figure}[t!]
    \centering
    \includegraphics[trim={100 0 200 0},clip,scale=0.4,page=8]{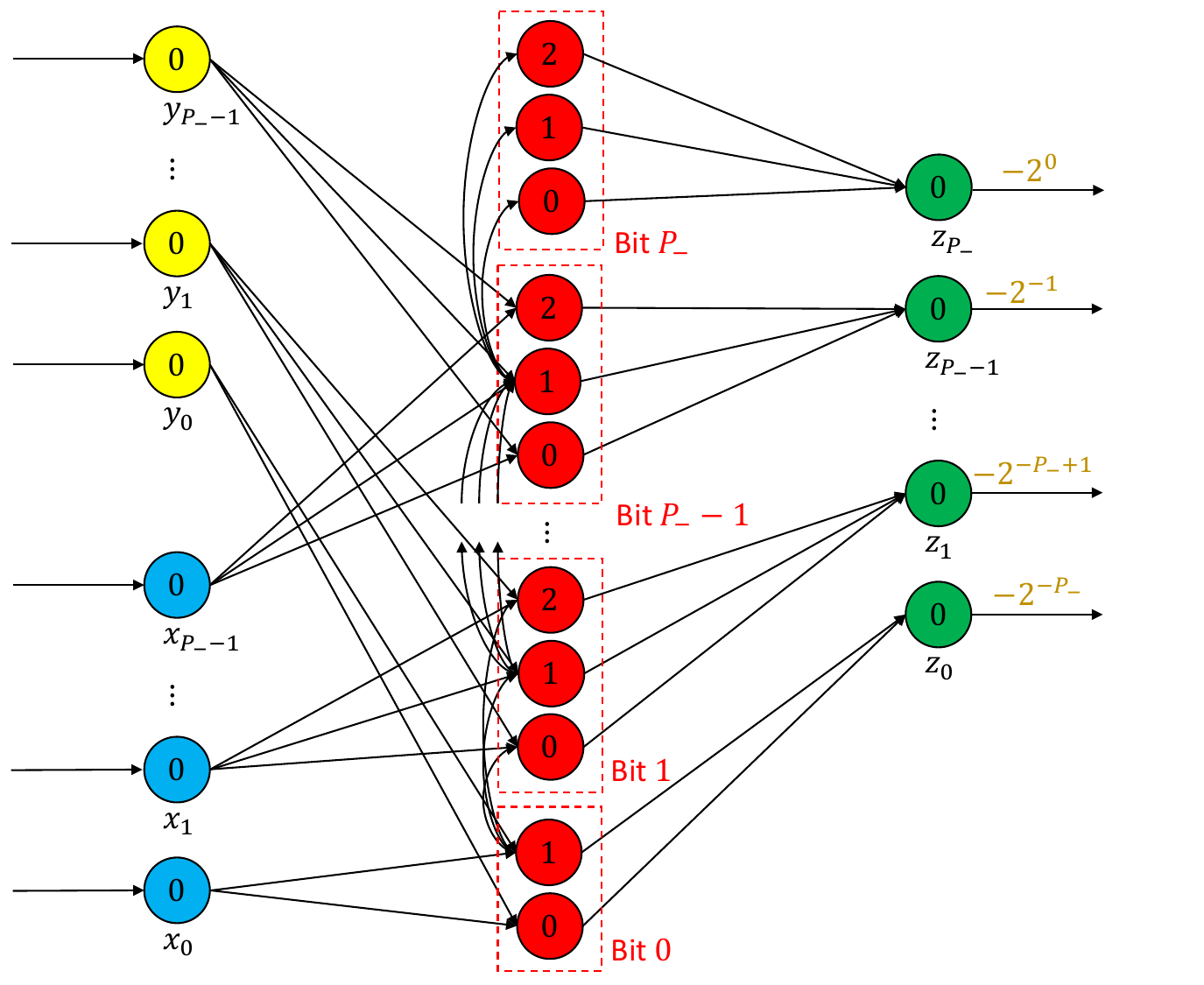}
    \caption{$P_-$ bit virtual neuron for encoding negative fractionals. Synapse parameters are omitted for brevity. Actual synapse parameters are assumed to be similar to the circuit shown in Figure \ref{fig:2-bit-virtual-neuron}.}
    \label{fig:negative-fractionals}
\end{figure}

Figure \ref{fig:negative-fractionals} shows the $P_-$ bit virtual neuron circuit for encoding negative fractionals.
This circuit is identical to Figure \ref{fig:negative-integers}, except the outgoing synapses have weights that are negatives of negative powers of two, i.e. $-2^0, -2^{-1}, -2^{-2}, \ldots$.

\subsection{Positive and Negative Rational Numbers}
\label{sub:virtual-neuron/rationals}

\begin{figure}[t!]
    \centering
    \includegraphics[trim={80 0 380 0},clip,scale=0.45,page=9]{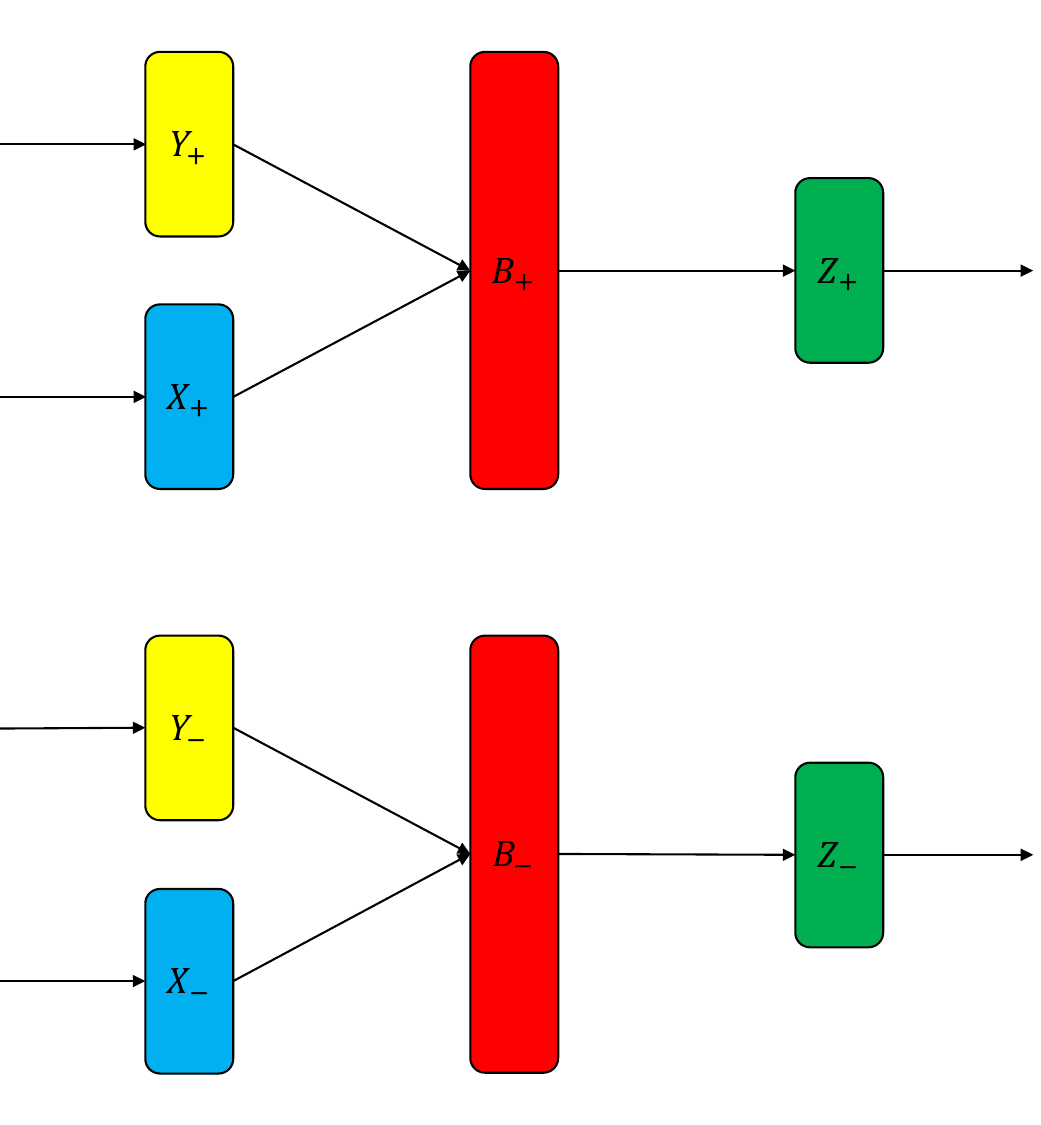}
    \caption{Encoding $P_+$ bit positive rationals and $P_-$ bit negative rationals using virtual neuron.}
    \label{fig:rationals}
\end{figure}

In this case, the virtual neuron operates on two $P_+ + P_-$ bit rational numbers $X$ and $Y$ as inputs.
These are shown in blue and yellow rectangles, which denote aggregation of respective neurons.
The positive precision $P_+$ is split between the positive integers and positive fractionals.
Similarly, negative precision is split between the negative integers and negative fractionals.
Notice that the positive part of the circuit (upper half) is completely independent from the negative part of the circuit (lower half).

\subsection{Computational Complexity}
\label{sub:virtual-neuron/theoretical-analysis}

\begin{table}[t!]
    \centering
    \caption{Neurons, synapses and time taken with increasing precision.}
    \begin{tabular}{@{}m{0.12\textwidth} m{0.07\textwidth} m{0.07\textwidth} m{0.12\textwidth}@{}}
        \noalign{\smallskip} \toprule \noalign{\smallskip}
        Number of bits of positive precision ($P_+$) & Number of neurons & Number of synapses & Time steps for virtual neuron operations  \\
        \noalign{\smallskip} \midrule \noalign{\smallskip}
        1 & 9 & 12 & 3 \\
        2 & 15 & 24 & 4 \\
        4 & 27 & 48 & 6 \\
        8 & 51 & 96 & 10 \\
        16 & 99 & 192 & 18 \\
        32 & 195 & 384 & 34 \\
        64 & 387 & 768 & 66 \\
        128 & 771 & 1536 & 130 \\
        \noalign{\smallskip} \bottomrule \noalign{\smallskip}
    \end{tabular}
    \label{tab:scalability}
\end{table}


For $P_+$ bit positive operations, we use $\mathcal{O}(P_+)$ neurons and synapses, and perform the virtual neuron operations in $\mathcal{O}(P_+)$ time steps.
Similarly, for $P_-$ bit negative operations, we use $\mathcal{O}(P_-)$ neurons and synapses, and perform the virtual neuron operations in $\mathcal{O}(P_-)$ time steps.
All in all, we use $\mathcal{O}(P_+ + P_-)$ neurons and synapses, and consume $\mathcal{O}(\max\{P_+, P_-\})$ time steps for the virtual neuron operations.

\begin{table}[]
\caption{Comparing Virtual Neuron to other neuromorphic encoding mechanisms for representing two N-bit numbers exactly.
}
\label{tab:neuromorphic_encoding}
\resizebox{\columnwidth}{!}{%
\begin{tabular}{@{}llllll@{}}
\toprule
\textbf{Metrics}                                                                                    & \textbf{Binning \cite{schuman2019non}} & \textbf{Rate \cite{schuman2019non}}   & \textbf{Time \cite{schuman2019non}} & \textbf{\begin{tabular}[c]{@{}l@{}}Virtual\\ Neuron\end{tabular}} & \textbf{\begin{tabular}[c]{@{}l@{}}IEEE-754\\ \cite{George2019,Dubey2020}\end{tabular}}                                               \\ \midrule
\begin{tabular}[c]{@{}l@{}}Time \\ Required \\ (big-$\mathcal{O}$)\end{tabular}                                 & $\mathcal{O}$(1)             & $\mathcal{O}$($2^N$)        & $\mathcal{O}$($2^N$)      & $\mathcal{O}$(1)                                                              & $\mathcal{O}$(1)                                                                                                                                                   \\ \midrule
\begin{tabular}[c]{@{}l@{}}\# Neurons\\ (big-$\mathcal{O}$)\end{tabular}                                        & $\mathcal{O}$($2^N$)         & $\mathcal{O}$(1)            & $\mathcal{O}$(1)          & $\mathcal{O}$(N)                                                              & \begin{tabular}[c]{@{}l@{}}Ensembles \\ of $N$ neurons \\ with dimension \\ and radius\\ properties$^1$\\ $\mathcal{O}$(N)$^2$\end{tabular}                               \\ \midrule
Accuracy                                                                                            & 100\%            & 100\%           & 100\%         & 100\%                                                             & \begin{tabular}[c]{@{}l@{}}\textgreater 90\% at 500 \\ Neurons per bit.\\ Encoded error \\ at 0 with \textgreater 300 \\ Neurons per bit.\end{tabular} \\ \midrule
\begin{tabular}[c]{@{}l@{}}Energy \\ Efficiency \\ (\# of spikes \\ in the worst case)\end{tabular} & $\mathcal{O}$(1) spikes       & $\mathcal{O}$($2^N$) spikes & $\mathcal{O}$(1) spikes   & \begin{tabular}[c]{@{}l@{}}$\mathcal{O}$(N) spikes; \\ N spikes\end{tabular}  & \begin{tabular}[c]{@{}l@{}}Energy results \\ unpublished\end{tabular}                                                                                  \\ \midrule
\begin{tabular}[c]{@{}l@{}}Energy \\ Efficiency \\ (\# spikes in \\ the average case)\end{tabular}   & $\mathcal{O}$(1) spikes       & $\mathcal{O}$($2^N$) spikes  & $\mathcal{O}$(1) spikes   & \begin{tabular}[c]{@{}l@{}}$\mathcal{O}$(N) spikes; \\ N/2 spikes\end{tabular}       & \begin{tabular}[c]{@{}l@{}}Energy results \\ unpublished\end{tabular}                                                                                  \\ \bottomrule
\end{tabular}
}
\scriptsize{
\\
$^1$Dimension refers to the number of values represented by the ensemble. (For a scalar quantity this is 1). Radius defines the range of values that can be represented by the ensemble. For the cited work dimension is 1 and radius is set to 2.\\
$^2$Authors only looked at IEEE floating point, so how the representation scales with numerical precision is unclear.
}
\end{table}

We validate these space and time complexities empirically for positive operations by increasing $P_+$.
The results of this analysis apply to negative operations as well.
We increase the positive precision from $1, 2, 4, \ldots, 128$ and count the number of neurons, synapses and time steps in each case.
The numerical results are presented in Table \ref{tab:scalability}.
From the table, we can conclude that we use $6P_+ + 3$ neurons, $12P_+$ synapses and $P_+ + 2$ time steps for virtual neuron operations.

\begin{table}[]
\caption{Comparing Virtual Neuron to other neuromorphic encoding mechanisms for adding two N-bit numbers.
}
\label{tab:neuromorphic_addition}
\resizebox{\columnwidth}{!}{%
\begin{tabular}{@{}lrrrr@{}}
\toprule
\textbf{Metrics}                                                         & \textbf{Binning \cite{schuman2019non}}        & \textbf{Rate Encoding \cite{schuman2019non}}  & \textbf{Virtual Neuron} & \textbf{\begin{tabular}[c]{@{}l@{}}IEEE 754 \cite{George2019}\end{tabular}}   \\
\midrule
Time to solution                                                         & $\mathcal{O}$(1)                    & $\mathcal{O}$($2^N$) & $\mathcal{O}$(N)                    & Constant                                                                  \\ \midrule
\# Neurons                                                               & $\mathcal{O}$($2^N$) & $\mathcal{O}$(1)                    & $\mathcal{O}$(N)                    & 3075                                                                         \\ \midrule
\# Synapses                                                              & $\mathcal{O}$($2^N$) &      $\mathcal{O}$(1)                & $\mathcal{O}$(N)                    & N/A                                                                           \\ \midrule
\begin{tabular}[c]{@{}l@{}}Energy \\ Efficiency \\ (\# spikes in \\ the worst case)\end{tabular} & $\mathcal{O}$($2^N$) & $\mathcal{O}$($2^N$) & $\mathcal{O}$(N)                    & N/A                                                                           \\ \midrule
\begin{tabular}[c]{@{}l@{}}Energy \\ Efficiency \\ (\# spikes in \\ the average case)\end{tabular}  & $\mathcal{O}$($2^N$) & $\mathcal{O}$($2^N$) & $\mathcal{O}$(N); N/2               & N/A                                                                           \\ \midrule
Accuracy                                                                 & 100\%$^1$                   & \begin{tabular}[c]{@{}l@{}}Depends on\\model type\end{tabular}             & 100\%                   & \begin{tabular}[c]{@{}l@{}}100\% with \\ 300 Neurons \\ per bit\end{tabular} \\ 
\bottomrule
\end{tabular}%
}
\scriptsize{
\\
$^1$Accuracy is bound by the synapse weight and accumulation accuracy.}
\end{table}

We can extend these time complexities to negative operations to conclude that they would require $6P_- + 3$ neurons, $12P_-$ synapses and $P_- + 2$ time steps.
This validates the space complexity as needing $\mathcal{O}(P_+ + P_-)$ neurons and synapses.
Since the positive and negative operations happen parallely, the overall time complexity of the circuit would stem from the larger of $P_+$ and $P_-$.
So, the overall time complexity is validated as $\mathcal{O}(\max\{P_+, P_-\})$.

Lastly, in computing the above space and time complexities, our inherent assumption is that the positive and negative precisions are variable.
However, we envision using the virtual neuron in settings where a neuromorphic computer has a fixed predetermined positive and negative precision.
This is similar to how the precision on our laptops and desktops is fixed to 32, 64 or 128 bits.
In such a scenario, $P_+$ and $P_-$ can be treated as constants.
Thus, the resulting space and time complexities for virtual neuron would all be $\mathcal{O}(1)$.


Table~\ref{tab:neuromorphic_encoding} presents a comparison of different neuromorphic encoding approaches in the literature with our approach using the virtual neuron. 
Since a neuromorphic computer consumes energy that is proportional to the number of spikes, we use the number of spikes in the worst and average case as an estimate for the energy usage of different neuromorphic approaches.
It can be seen that across different comparison metrics such as network size, or number of spikes, the virtual neuron scales linearly with the bit-precision $N$, while giving the exact representation of the input number.
Other approaches take either exponential space (Binning), or exponential time (Rate Encoding), or are unable to represent rational numbers exactly (IEEE 754).
Table~\ref{tab:neuromorphic_addition} presents the comparison of computational complexity for performing addition with two N-bit numbers under different neuromorphic encoding schemes. Here we do not include temporal encoding scheme because under such a simple approach, binary spikes occurring at different time instances cannot be added in an exact manner by spiking neurons.
While the virtual neuron can perform the addition operation in linear time steps and using linear number of neurons, synapses and energy (as estimated by the spiking efficiency), other approaches use either exponential time or exponential space or consume exponential amount of energy for their operations.

\section{Implementation Details}
\label{sec:implementation-details}

We implemented the virtual neuron in Python using the NEST simulator.
The hardware on which the simulations were run was a MacBook Pro having a 2.3 GHz Quad-Core Intel Core i7 processor and 32 GB 3733 MHz LPDDR4X memory.
We wrote a \texttt{VirtualNeuron} class, whose constructor took a list-like object of length 4 as the precision vector.
The elements of this vector corresponded to number of bits for positive integers, positive fractionals, negative integers and negative fractionals.
We then computed the positive precision as the sum of the first two elements of the precision vector, and the negative precision as the sum of the third and fourth elements of the precision vector.

We then created all the neurons and set their parameters correctly.
We used the \texttt{iaf\_psc\_delta} neuron model.
All neurons had an internal state of $-1.0$.
In NEST, the internal state corresponds to the voltage of the membrane potential (\texttt{V\_m}) parameter.
All neurons had a leak of $10^{-6}$, which is a good approximation to $0$ leak that we require in our circuits.
In NEST, the leak corresponds to the \texttt{tau\_m} neuron parameter.
All neurons except the bit neurons (red neurons) had a neuron threshold of $0$.
The group of bit neurons corresponding to the least significant bit in both the positive and negative parts of the circuit had only two neurons with thresholds $0$ and $1$.
All other groups of bit neurons had three neurons with thresholds $0$, $1$ and $2$ respectively.
There were $P_+$ ($P_-$) such groups in the positive (negative) part of the circuit, making a total of $P_+ + 1$ ($P_- + 1$) groups of bit neurons, corresponding to the $P_+ + 1$ ($P_- + 1$) output bits in the positive (negative) parts of the circuit.

After the neurons were created, we setup the synapses.
Firstly, synapses between the positive (negative) incoming neurons and positive (negative) bit neurons were created.
These synapses had synaptic weights as $1.0$ and synaptic delays as \texttt{i+1}, where \texttt{i} ranges from $0$ to $P_+$ ($P_-$).
Secondly, we setup the carry synapses between the consecutive groups of positive (negative) bit neuron groups.
The carry synapses go from the bit neuron having a threshold of $1$ in the $i^{\text{th}}$ group to all neurons in the $(i+1)^{\text{th}}$ group, where \texttt{i} goes from $0$ to $P_+$ ($P_-$).
The carry synapses had both weights and delays as $1.0$.
Finally, we setup synapses from groups of positive (negative) bit neurons to their corresponding outgoing neurons.
The synapses coming from bit neurons with thresholds $0$ and $2$ had weights $1.0$, whereas those coming from bit neurons with thresholds $1$ had weights of $-1.0$.
Furthermore, these syapses in the positive (negative) part of the circuit had delays given by $\max\{P_+, P_-\} - i + 1$ for \texttt{i} ranging from $0$ to $P_+$ ($P_-$).
We also wrote a function \texttt{connect\_virtual\_neurons(A, B, C)}, that connects three virtual neurons \texttt{A}, \texttt{B} and \texttt{C} such that \texttt{A} and \texttt{B} serve as inputs to \texttt{C}.
The weights and delays on these synapses were all $1.0$.

\section{Testing Results}
\label{sec:testing-results}

We tested our implementation of the virtual neuron on 8, 16 and 32 bit rational numbers.
The precision vectors fed to the class constructors in each of these cases were $[2,2,2,2]$, $[4,4,4,4]$, and $[8,8,8,8]$ respectively.
We connected three virtual neurons using the \texttt{connect\_virtual\_neurons} function described above.
Next, we generated two numbers within the appropriate precision by generating spikes through the \texttt{spike\_generator} in NEST, and then sent these spikes to the input virtual neurons \texttt{A} and \texttt{B}.
We let the simulation run for a time long enough so that we receive an output from virtual neuron \texttt{C}.
Lastly, we checked if output received from \texttt{C} was indeed the sum of numbers sent to \texttt{A} and \texttt{B}.


\subsection{8 Bit Virtual Neuron}
\label{sub:testing/8-bit}

\begin{table*}[]
    \centering
    \caption{Testing virtual neuron on 8-bit rational numbers. Precision is [2, 2, 2, 2] for positive integer, positive fraction, negative integer and negative fraction respectively.}
    \begin{tabular}{m{0.06\textwidth} m{0.06\textwidth} m{0.06\textwidth} m{0.06\textwidth} | m{0.06\textwidth} m{0.06\textwidth} m{0.06\textwidth} m{0.06\textwidth} | m{0.06\textwidth} m{0.06\textwidth} m{0.06\textwidth} m{0.06\textwidth}}
        \noalign{\smallskip} \hline \noalign{\smallskip}
        \multicolumn{2}{c}{$X_+$} & \multicolumn{2}{c|}{$X_-$} & \multicolumn{2}{c}{$Y_+$} & \multicolumn{2}{c|}{$Y_-$} & \multicolumn{2}{c}{$Z_+$} & \multicolumn{2}{c}{$Z_-$} \\
        Decimal & Binary & Decimal & Binary & Decimal & Binary & Decimal & Binary & Decimal & Binary & Decimal & Binary \\
        \noalign{\smallskip} \hline \noalign{\smallskip}
        0.75 & 0011 & -2.75 & 1011 & 1.0  & 0100 & -2.5  & 1010 & 1.75 & 00111 & -5.25 & 10101 \\
        2.5  & 1010 & -3.75 & 1111 & 1.75 & 0111 & -0.25 & 0001 & 4.25 & 10001 & -4.0  & 10000 \\
        0.25 & 0001 & -2.75 & 1011 & 2.75 & 1011 & 0.0   & 0000 & 3.0  & 01100 & -2.75 & 01011 \\
        3.5  & 1110 & -2.5  & 1010 & 3.5  & 1110 & -0.25 & 0001 & 7.0  & 11100 & -2.75 & 01011 \\
        3.0  & 1100 & 0.0   & 0000 & 3.25 & 1101 & -1.0  & 0100 & 6.25 & 11001 & -1.0  & 00100 \\
        \noalign{\smallskip} \hline \noalign{\smallskip}
    \end{tabular}
    \label{tab:8-bit-testing}
\end{table*}

For the 8 bit case, we tested all permutations of the input numbers---a total of $65,536$ cases.
A randomly selected sample of results are shown in Table \ref{tab:8-bit-testing}.
It can be seen clearly that $X_+ + Y_+ = Z_+$, and $X_- + Y_- = Z_-$ for all rows.
The binary representations of $X$ and $Y$ were fed to the input neurons and the binary representation of $Z$ was received as the output of the circuit.

\subsection{16 Bit Virtual Neuron}
\label{sub:testing/16-bit}

\begin{table*}[]
    \centering
    \caption{Testing virtual neuron on 16-bit rational numbers. Precision is [4, 4, 4, 4] for positive integer, positive fraction, negative integer and negative fraction respectively.}
    \begin{tabular}{m{0.06\textwidth} m{0.06\textwidth} m{0.06\textwidth} m{0.06\textwidth} | m{0.06\textwidth} m{0.06\textwidth} m{0.06\textwidth} m{0.06\textwidth} | m{0.06\textwidth} m{0.06\textwidth} m{0.06\textwidth} m{0.06\textwidth}}
        \noalign{\smallskip} \hline \noalign{\smallskip}
        \multicolumn{2}{c}{$X_+$} & \multicolumn{2}{c|}{$X_-$} & \multicolumn{2}{c}{$Y_+$} & \multicolumn{2}{c|}{$Y_-$} & \multicolumn{2}{c}{$Z_+$} & \multicolumn{2}{c}{$Z_-$} \\
        Decimal & Binary & Decimal & Binary & Decimal & Binary & Decimal & Binary & Decimal & Binary & Decimal & Binary \\
        \noalign{\smallskip} \hline \noalign{\smallskip}
        2.5625  & 00101001 & -11.375  & 10110110 & 13.3125 & 11010101 & -6.75   & 01101100 & 15.875  & 011111110 & -18.125  & 100100010 \\
        2.3125  & 00100101 & -13.9375 & 11011111 & 11.375  & 10110110 & -9.3125 & 10010101 & 13.6875 & 011011011 & -23.25   & 101110100 \\
        15.875  & 11111110 & -2.9375  & 00101111 & 1.5625  & 00011001 & -4.6875 & 01001011 & 17.4375 & 100010111 & -7.625   & 001111010 \\
        8.625   & 10001010 & -10.1875 & 10100011 & 8.9375  & 10001111 & -1.625  & 00011010 & 17.5625 & 100011001 & -11.8125 & 010111101 \\
        14.6875 & 11101011 & -10.625  & 10101010 & 11.625  & 10111010 & -11.875 & 10111110 & 26.3125 & 110100101 & -22.5    & 101101000 \\
        \noalign{\smallskip} \hline \noalign{\smallskip}
    \end{tabular}
    \label{tab:16-bit-testing}
\end{table*}

The results from the 16 bit testing are shown in Table \ref{tab:16-bit-testing}.
In this case, we tested $100,000$ permutations of inputs, generated uniformly at random.
A snippet of the results are shown in Table \ref{tab:16-bit-testing}.
One can infer that the virtual neuron is mimicking an artificial neuron having an identity activation function, and that we are able to encode positive and negative rational numbers on a neuromorphic computer using this approach.

\subsection{32 Bit Virtual Neuron}
\label{sub:testing/32-bit}

\begin{table*}[]
    \centering
    \caption{Testing virtual neuron on 32-bit rational numbers. Precision is [8, 8, 8, 8] for positive integer, positive fraction, negative integer and negative fraction respectively.}
    \begin{tabular}{m{0.12\textwidth} m{0.12\textwidth} | m{0.12\textwidth} m{0.12\textwidth} | m{0.12\textwidth} m{0.12\textwidth}}
        \noalign{\smallskip} \hline \noalign{\smallskip}
        $X_+$ & $X_-$ & $Y_+$ & $Y_-$ & $Z_+$ & $Z_-$ \\
        \noalign{\smallskip} \hline \noalign{\smallskip}
        212.56640625 & -203.421875 & 218.7265625 & -98.91796875 & 431.29296875 & -302.33984375 \\
        1.375 & -4.36328125 & 184.94921875 & -92.73046875 & 186.32421875 & -97.09375 \\
        254.3359375 & -134.390625 & 48.87109375 & -211.43359375 & 303.20703125 & -345.82421875 \\
        44.203125 & -231.0703125 & 177.1171875 & -207.06640625 & 221.3203125 & -438.13671875 \\
        143.6171875 & -8.1171875 & 214.41796875 & -224.01953125 & 358.03515625 & -232.13671875 \\
        \noalign{\smallskip} \hline \noalign{\smallskip}
    \end{tabular}
    \label{tab:32-bit-testing}
\end{table*}

For the 32 bit case, we generated $100,000$ permutations of inputs uniformly at random.
Five randomly selected permutations are presented in Table \ref{tab:32-bit-testing}.
Once again, it can be concluded that the virtual neuron is successfully able to encode and add rational numbers on neuromorphic computers and scales linearly with the positive and negative precisions.

\subsection{Caspian and Hardware Testing}
We also implemented and tested the 16-bit virtual neuron using the Caspian simulator and \ucaspian\ digital FPGA hardware \cite{10.1145/3381755.3381764}. Since \ucaspian\ does not implement synaptic delay, but instead implements axonal delay, the virtual neuron implementation was adjusted to use axonal delay instead of synaptic delay. The rest of the structure is the same as the NEST virtual neuron implementation, this means that the neuron and synapse counts and network time steps to solution are the same as in the NEST implementation.

\ucaspian\ is a digital neuromorphic processor implementation using an FPGA, the processor is event-based and processes all the spikes that occur at one time step before moving to the next time step.
Time multiplexing of neurons is used to reduce the size of the design.
\ucaspian\ is intentionally designed targeting the small and low-power iCE40 UP5k FPGA. Because of this, \ucaspian\ only supports up to 256 neurons and 4096 synapses.
This is enough to support up to a 32-bit virtual neuron adder; however, to include room for the input and output neurons, we tested with the 16-bit virtual neuron.
Since \ucaspian\ run time depends on activity, we ran 1,000 permutations of inputs selected uniformly at random on the \ucaspian\ simulator and hardware, and monitored the total number of spikes and the number of cycles used by the processor.
\ucaspian\ has a behaviourally accurate software simulator and the hardware design can be emulated in Verilator or run on the FPGA.
In this case we used the UPdruino V3 as the FPGA board.

Over the 1,000 runs, the simulator reported 73,159 total spikes for an average of $73$ spikes per test case.
Using Verilator, the 1,000 test cases finished in $\sim$ 5,000,000 clock cycles.
Where $\sim$ 7,000 cycles where used to load the virtual neuron network and $\sim$ 5,000 cycles are used per test case.
Since the processor runs at 25 MHz, the total runtime without the overheads from communication with the host computer is $\sim 0.21$s for all the test cases.
When we ran the test using the UPduino FPGA, the total time was $\sim$ 400s.
One main culprit for this slowdown is the 3 MBaud UART connection between the host and the FPGA.
While running on hardware, over $99.9\%$ of the execution time was spent in overhead and communication. This result highlights the great benefit of using the virtual neuron to perform addition on the SNN system instead of moving the data to a separate processor to perform the addition.
The results from the hardware evaluation are tabulated in Table \ref{tbl:hardware_eval} and a summary of the \ucaspian\ processor cycles from the experiment are in Table \ref{tbl:hardware_cycles}.

\subsection{mrDANNA Power Estimate}
With neuromorphic application-specific integrated circuits, the power required for a particular network execution can be estimated based on the energy required for active and idle neurons and synapses for the duration of the execution.
To estimate the power of the virtual neuron design, we used the same method and energy-per-spike values as reported in \cite{Chakma2018a} for the mrDANNA mixed-signal memristor-based neuromorphic processor. Using the same number of spikes, neurons, and synapses as reported in the \ucaspian\ simulation, we estimate that a mrDANNA hardware implementation would use $\sim 23$ nJ for the average test case run and around $\sim 23$ mW for continuous operation.

\begin{table}[]
\centering
\caption{Summary of Hardware Evaluation}
\label{tbl:hardware_eval}
\begin{tabular}{@{}lrrr@{}}
\toprule
Method             & \begin{tabular}[c]{@{}r@{}}Execution Time \\ of Processor\end{tabular} & \begin{tabular}[c]{@{}r@{}}Wall Time \\ of Evaluation\end{tabular} & Power    \\ \midrule
µCaspian Hardware  & 0.21 s                                                                 & 400 s                                                              &          \\
µCaspian Simulator & N/A                                                                    & 747 ms                                                             &          \\
mrDANNA            & 1 µs @ 20MHz                                                           & N/A                                                                & 23.04 mW \\ \bottomrule
\end{tabular}
\end{table}

\begin{table}[]
\centering
\caption{\ucaspian\ Cycles summary}
\label{tbl:hardware_cycles}
\begin{tabular}{@{}lr@{~}l@{}} \toprule
\multicolumn{3}{c}{\textbf{Entire   Test}}                \\ \midrule
Clock Cycles     & 5,000,000 &                            \\
Total Time       & 0.21                          & s      \\
Per Test Average & 0.21                          & ms     \\ \bottomrule
\end{tabular}

\vspace*{.1in}

\begin{tabular}{@{}lr@{~}l@{}} \toprule
\multicolumn{3}{c}{\textbf{Single Test Time}}             \\ \midrule
Clock Cycles     & 5,000                         &        \\
Time             & 0.21                          & ms     \\ \bottomrule
\end{tabular}
\hspace{.1in}
\begin{tabular}{@{}lr@{~}l@{}} \toprule
\multicolumn{3}{c}{\textbf{Network Load Time}}            \\ \midrule
Clock Cycles     & 7,000                         &        \\
Time             & 0.29                          & ms     \\ \bottomrule
\end{tabular}
\end{table}

\color{black}

\section{Applications}
\label{sec:applications}

In this section, we look at five functions where virtual neuron is used: constant function, successor function, predecessor function, multiply by $-1$ function, and $N$-neuron addition.

\subsection{Constant Function}
\label{sub:applications/constant}

\begin{figure}[t!]
    \centering
    \includegraphics[trim={200 160 200 150},clip,scale=0.5,page=10]{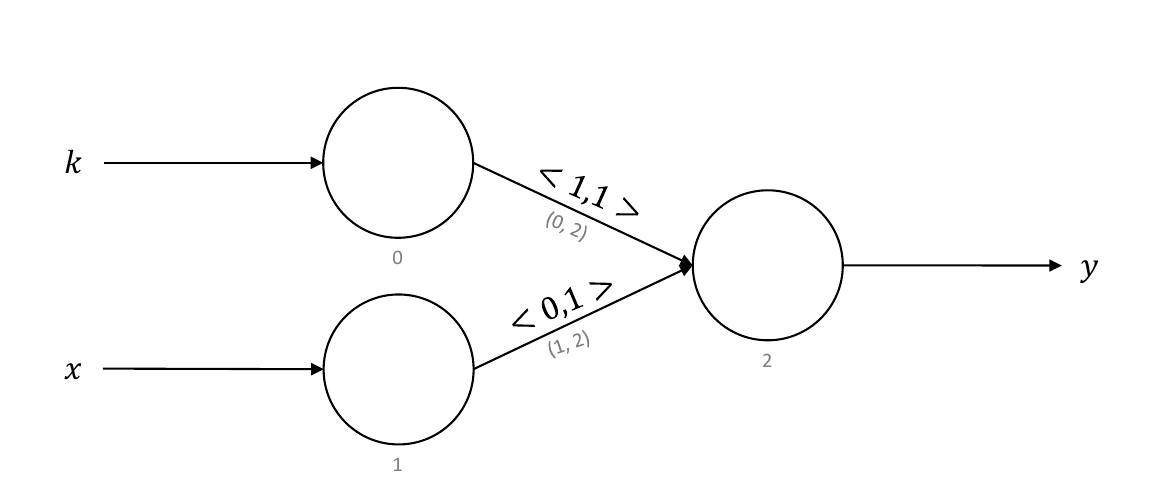}
    \caption{Neuromorphic circuit for constant function}
    \label{fig:constant-function}
\end{figure}

For a natural number $x$, the constant function returns a constant natural number $k$. It is defined as:
\begin{align}
    C_k(x) := k \label{eq:constant-function-definition}
\end{align}

Figure \ref{fig:constant-function} shows the neuromorphic circuit that computes the constant function.
It has been adapted from \cite{date2021neuromorphic} to work with the virtual neuron.
Each neuron in this circuit is a virtual neuron.
Inputs $k$ and $x$ are fed to the input neurons \texttt{0} and \texttt{1}, and the output is produced at neuron \texttt{2}.
Synapses going from neuron \texttt{0} to neuron \texttt{2} have weights of $1$, while those going from neuron \texttt{1} to neuron \texttt{2} have weights of $0$.
The constant function is one of the $\mu$-recursive functions.
$\mu$-recursion is a model of computation that is equivalent to the Turing machine.
In order to prove that a computing platform is Turing-complete, it suffices to prove that it can execute all the $\mu$-recursive functions.
In that light, being able to implement the constant function is a step towards empirically showing that neuromorphic computing is Turing-complete.
We implemented the constant function circuit in NEST and tested it with $16$ bit natural numbers.
We were able to accurately execute the constant function using virtual neurons.

\subsection{Successor Function}
\label{sub:applications/successor}

\begin{figure}[t!]
    \centering
    \includegraphics[trim={200 160 200 150},clip,scale=0.5,page=11]{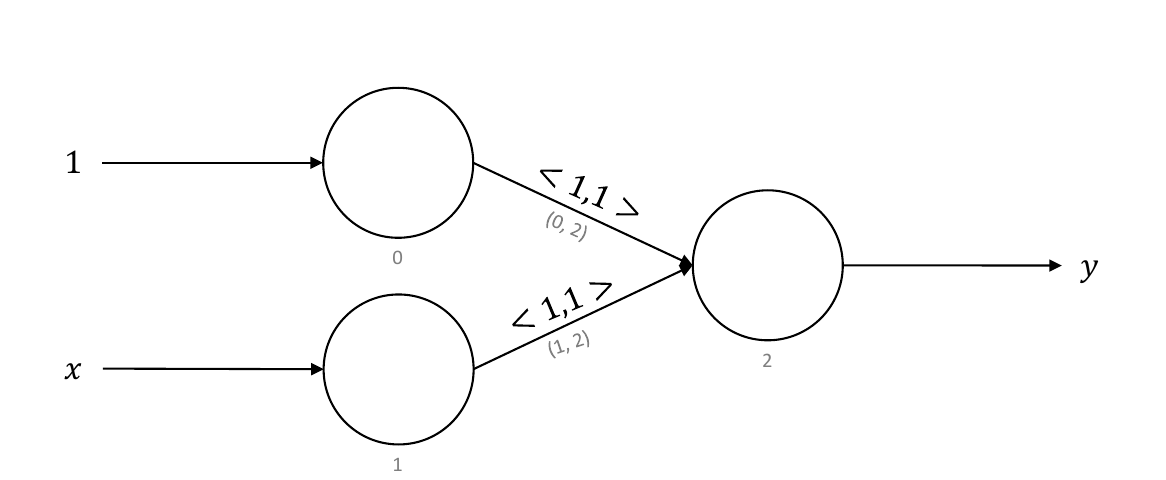}
    \caption{Neuromorphic circuit for successor function}
    \label{fig:successor-function}
\end{figure}

For a natural number $x$, the successor function returns $x+1$.
The successor of $0$ is defined as $1$.
The successor function is defined as:
\begin{align}
    S(x) := x + 1 \label{eq:successor-function-definition}
\end{align}

Figure \ref{fig:successor-function} shows the successor function.
It too has been adapted from \cite{date2021neuromorphic} and is another $\mu$-recursive function.
It is similar to the constant function with a couple of differences.
Neuron \texttt{0} is fed an input of $1$ and synapse \texttt{(1,2)} has a weight of $1$.
We implemented the successor function using three virtual neurons and tested it on 16-bit numbers.
Our implementation was able to execute the successor function successfully.

\subsection{Predecessor Function}
\label{sub:applications/predecessor}

\begin{figure}[t!]
    \centering
    \includegraphics[trim={200 160 200 150},clip,scale=0.5,page=12]{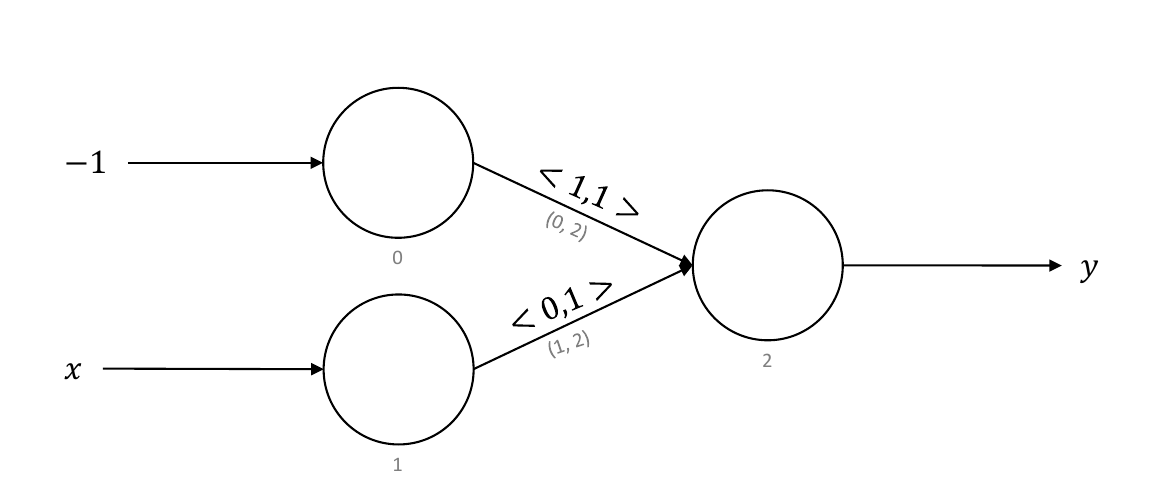}
    \caption{Neuromorphic circuit for predecessor function}
    \label{fig:predecessor-function}
\end{figure}

For a natural number $x$, the predecessor function returns $x-1$.
The predecessor function is defined as:
\begin{align}
    S(x) := x - 1 \label{eq:predecessor-function-definition}
\end{align}

Figure \ref{fig:predecessor-function} shows the predecessor function.
It is similar to the successor function with just one change.
We feed an input of $-1$ to neuron \texttt{0} as opposed to $1$.
We implemented the predecessor function using three virtual neurons and tested it on $16$ bit numbers.
We were able to execute the predecessor function successfully using three virtual neurons in NEST.

\subsection{Multiply by -1}
\label{sub:applications/multiply-by--1}

\begin{figure}[t!]
    \centering
    \includegraphics[trim={300 150 300 150},clip,scale=0.55,page=13]{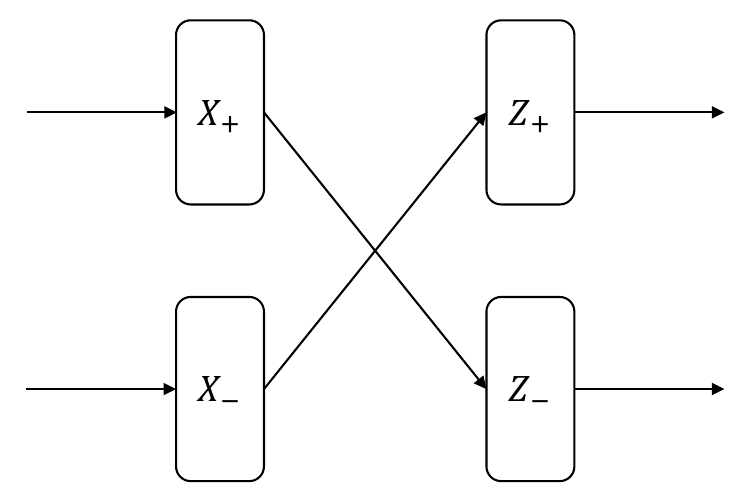}
    \caption{Neuromorphic circuit to multiply a number by $-1$.}
    \label{fig:mul-by-neg-one-function}
\end{figure}

For a rational number $x$ this function returns $-x$.
Figure \ref{fig:mul-by-neg-one-function} shows the multiply by $-1$ function.
It takes a rational number encoded in virtual neuron $X$.
In the figure, we use $X_+$ to denote the positive part of $X$ and $X_-$ to denote negative parts of $X$.
In this function, we assume that the number of positive and negative precision bits are equal.
Under this assumption, we simply swap the positive and negative parts of $X$ to return a number $Z$ encoded as a virtual neuron.
Since $Z_+$ equals $X_-$ and $Z_-$ equals $X_+$, this function returns the negative of a number fed as the input.
We implemented this function on $16$ bit numbers and found that our virtual neuron-based implementation was able to execute the function successfully.

\subsection{N-Neuron Addition}
\label{sub:applications/multi-neuron-addition}

\begin{figure}[t!]
    \centering
    \includegraphics[trim={100 0 100 0},clip,scale=0.32,page=14]{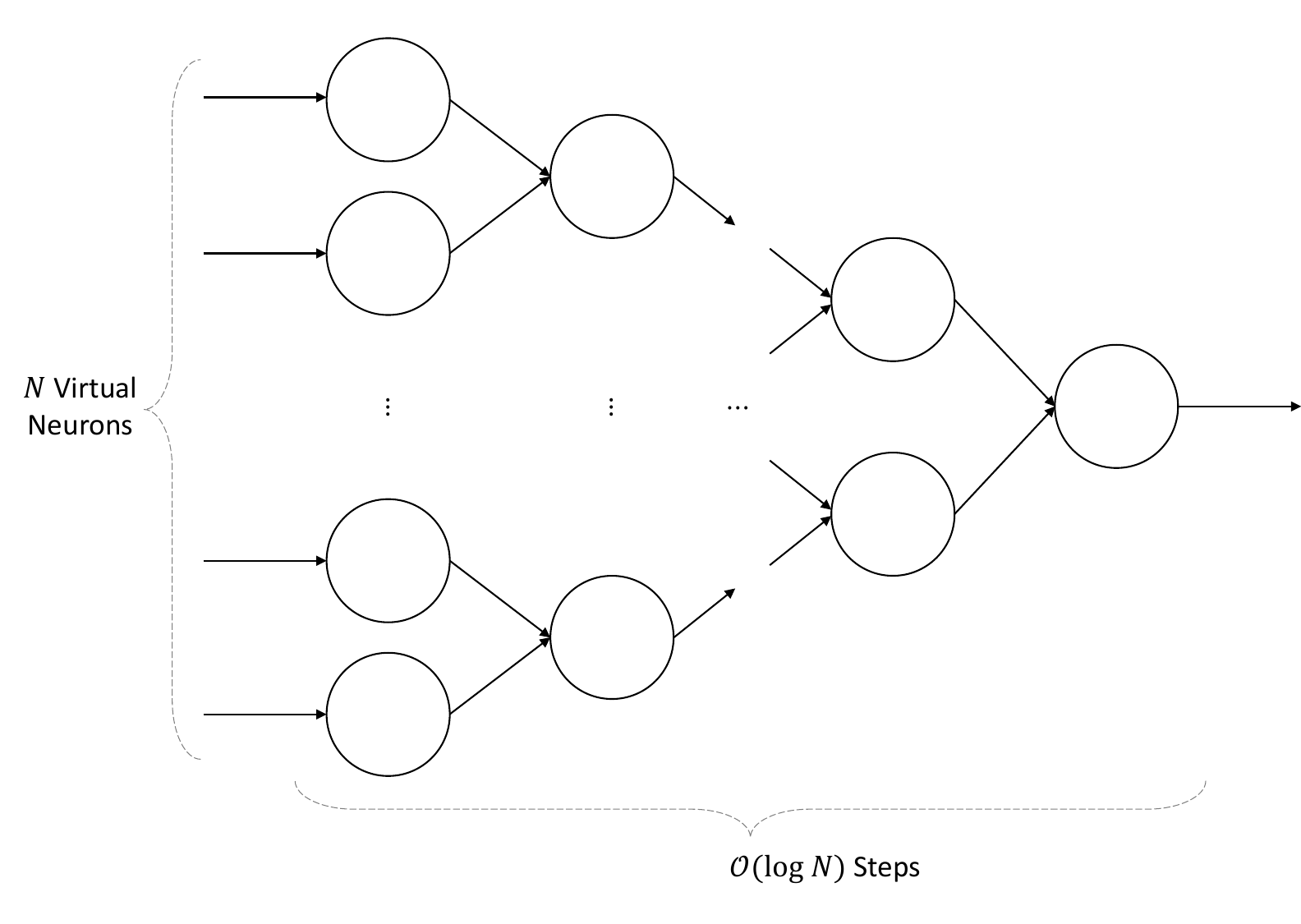}
    \caption{Neuromorphic circuit to add $N$ virtual neurons.}
    \label{fig:n-neuron-addition}
\end{figure}

The last application of virtual neuron that we want to highlight is the $N$-neuron addition.
The figure for this application is shown in Figure \ref{fig:n-neuron-addition}, where we would like to add $N$ virtual neurons given as inputs.
The addition is performed by successfully connecting pairs of input virtual neurons to a layer of virtual neurons, which in turn serve as inputs to the next layer.
This method uses $\mathcal{O}(N)$ virtual neurons and synapses and runs in $\mathcal{O}(\log N)$ time steps.
We implemented the $N$-neuron addition circuit in NEST and tested it on $16$ bit numbers.
This implementation was successfully able to add $N$ virtual neurons in $\mathcal{O}(\log N)$ time.

\section{Discussion}

In this paper, we proposed the virtual neuron as a mechanism for encoding as well as adding positive and negative rational numbers.
Our work is a stepping stone towards a broader class of neuromorphic computing algorithms, that would enable us to perform general-purpose computations on neuromorphic computers.
In this paper, we also measured the time, space, and energy required for virtual neuron operations and showed that it takes around 23 mW of power. 
In addition to a low operational power, there will be great savings in performing the operation withing the spiking array, without the need to spend energy sending the data to an external processor to perform the operation.
Although it is out of the scope of this paper, the virtual neuron is a vital component to enabling composing of sub-networks to scale-up neuromorphic algorithms, and it is also a vital component to support within network encoding and decoding capabilities.
We would like to address these areas as part of our future work.

The virtual neuron can be viewed as a tool that enables bit-level precision as well as variable precision on neuromorphic computers.
We also expect the virtual neuron to be used in neuromorphic compilers that can compile high-level neuromorphic algorithms down to neurons and synapses, which can be deployed onto neuromorhpic hardware directly.
Another potential use case of the virtual neuron is to encode and perform operations on extremely large numbers (containing thousands of bits), as required in many cryptography applications.
The virtual neuron would enable us to perform these large number operations in an energy efficient manner on neuromorphic computers.
In an edge computing scenario, current neuromorphic computers allow us to perform machine learning tasks using spiking neural networks in an energy-efficient manner.
However, if an application requires performing general-purpose operations on data (for instance, pre- or post-processing of data using arithmetic, logical and relational operations), we resort to conventional computers (CPUs and GPUs), which incurs significant communication cost.
With approaches such as virtual neuron that serve as the building blocks of general-purpose computing on neuromorphic computers, we could potentially perform all these operations on the neuromorphic computer itself without having to communicate to the CPU/GPU and bypassing the need to transfer data back and forth from the CPU/GPU.

It is worth noting that the ultimate goal of a neuromorphic computer is \textit{not} to perform these sorts of operations.  However, many applications for which a neuromorphic system might be used (for classification, anomaly detection, control, etc.) may require these sorts of calculations as a pre- or post-processing step for the neuromorphic system. For a continually operating neuromorphic system, for example in a control application, these sorts of calculations may be required \textit{between} neuromorphic calculations.  If these computations can take place \textit{on} the neuromorphic computer, then it will alleviate communication costs and data movement to and from the neuromorphic system.  As such, even if the computations described above are not as efficient as those on a traditional processor, it is likely that the data movement costs to and from the traditional processor will overwhelm the energy efficiency benefits gained from moving the computation back to a traditional processor.

Just like with IEEE standard data types, real numbers such as $\sqrt{2}, \pi$ and $e$ cannot be directly encoded without infinite bits of precision. Therefore the bits of precision used in the virtual neuron encoding can be chosen based on the accuracy of the approximation of the real number required.
Lastly, the applications demonstrated in Section \ref{sec:applications} might seem simple, but they are critical building blocks for any general-purpose computations that can be performed on a neuromorphic computer.
Complex general-purpose compute tasks can be broken down into the simplest of operations defined by these functions.
We would like to reiterate that the goal of this paper was to present the idea of the virtual neuron and demonstrate its performance on physical and simulated neuromorphic hardware.
The demonstration on the applications mentioned in Section \ref{sec:applications} is to give the reader an idea of how the virtual neuron can be used.

\color{black}

\section{Conclusion}
\label{sec:conclusion}

Neuromorphic computing is an extremely promising paradigm for energy efficient computing in the future.
It holds tremendous promise to drastically reduce the carbon footprint of computing.
While traditional applications of neuromorphic computing primarily relied on SNN-based machine learning, more recent applications in graph algorithms, autonomous racing and linear algebra show that neuromorphic computing might be capable of much more than just SNNs.
Neuromorphic computing has shown to be Turing-complete, and thus, capable of all general-purpose computation.
A key step to realize the full potential of general-purpose neuromorphic computing is to devise effective mechanisms of encoding numbers.
The current mechanisms for encoding numbers on neuromorphic computers are limited to Boolean numbers or natural numbers.
But even these mechanisms are limited to the specific applications for which they were developed, and are not suitable for general-purpose computation.
Moreover, some of these methods result in loss of data due to excessive discretization and/or do not preserve addition.

In this work, we presented the virtual neuron as a mechanism for encoding positive and negative integer and rational numbers.
We implemented the virtual neuron in the NEST simulator and tested it on 8, 16 and 32 bit rational numbers.
We theoretically compared the computational complexity of the virtual neuron to other neuromorphic encoding mechanisms.
Next, we tested the virtual neuron on neuromorphic hardware and presented its time, space and power metrics.
Lastly, we demonstrated the usability of the virtual neuron by using it in five applications, that would be crucial for general-purpose neuromorphic computing.
We were able to show that the virtual neuron is an efficient mechanism for encoding rational numbers.
Furthermore, we also showed that the virtual neuron can mimic the artificial neuron with an identity activation function.
In our future work, we would like to explore general-purpose neuromorphic algorithms and applications using virtual neurons.

\bibliographystyle{IEEEtran}
\bibliography{references}

\end{document}